# Using Clinical Narratives and Structured Data To Identify Distant Recurrences in Breast Cancer

Zexian Zeng
Preventive Medicine
Northwestern University
Feinberg School of Medicine
Chicago, IL, USA
zexian.zeng@northwestern.edu

Ankita Roy
Surgery
Northwestern University
Feinberg School of Medicine
Chicago, IL, USA
ankita.roy@northwestern.edu

Xiaoyu Li
Social and Behavioral Sciences
Harvard T.H. Chan School of Public Health
Boston, MA, USA
xil288@mail.harvard.edu

Sasa Espino
Surgery
Northwestern University
Feinberg School of Medicine
Chicago, IL, USA
sasa-grae.espino@northwestern.edu

Susan Clare
Surgery
Northwestern University
Feinberg School of Medicine
Chicago, IL, USA
susan.clare@northwestern.edu

Seema Khan
Surgery
Northwestern University
Feinberg School of Medicine
Chicago, IL, USA
s-khan2@northwestern.edu

Yuan Luo (corresponding)
Preventive Medicine
Northwestern University
Feinberg School of Medicine
Chicago, IL, USA
yuan.luo@northwestern.edu

*Abstract*—Accurately identifying distant recurrences in breast cancer from the Electronic Health Records (EHR) is important for both clinical care and secondary analysis. Although multiple applications have been developed for computational phenotyping in breast cancer, distant recurrence identification still relies heavily on manual chart review. In this study, we aim to develop a model that identifies distant recurrences in breast cancer using clinical narratives and structured data from EHR. We apply MetaMap to extract features from clinical narratives and also retrieve structured clinical data from EHR. Using these features, we train a support vector machine model to identify distant recurrences in breast cancer patients. We train the model using 1,396 double-annotated subjects and validate the model using 599 double-annotated subjects. In addition, we validate the model on a set of 4,904 single-annotated subjects as a generalization test. We obtained a high area under curve (AUC) score of 0.92 (SD=0.01) in the cross-validation using the training dataset, then obtained AUC scores of 0.95 and 0.93 in the held-out test and generalization test using 599 and 4,904 samples respectively. Our model can accurately and efficiently identify distant recurrences in breast cancer by combining features extracted from unstructured clinical narratives and structured clinical data.

## I. Introduction

Distant recurrences are defined as metastasis of the primary breast tumor to lymph nodes or organs beyond the loco-regional pathological field. Nodes located within the loco-regional field include ipsilateral axillary, ipsilateral internal mammary, supraclavicular, and intramammary lymph nodes [1]. Distant lymph nodes beyond the loco-regional field include cervical, contralateral axillary, and contralateral internal mammary lymph nodes. The most common sites of metastasis to organs are the bone, brain, lung, and liver [1]. It is important to distinguish between local and distant recurrences for several reasons: the categorization informs treatment decision-making and directs studies analyzing outcomes of local versus distant recurrences. Most importantly, the 10-year survival rates are much lower for distant recurrences as compared to local recurrences (56% after an isolated local recurrence as opposed to 9% after distant metastasis) [2]. The delineation can be an important prognostic marker for mortality.

The emerging cancer prognosis research has directed efforts towards identifying distant cancer recurrence events accurately and efficiently. The National Program of Cancer Registries (NPCR) was launched to capture cancer patient information and one of its major tasks is to capture disease prognosis status for each cancer patient. However, many tumor registries fail to accurately identify cancer recurrences due to the significant human effort required for data maintenance [3, 4]. Manual chart review is one of the traditional methods used to identify breast cancer recurrences. Unfortunately, chart review is a time-consuming and costly process. It limits the number of samples available for research

IEEE International Conference on Healthcare Informatics 2018 (ACCEPTED)and is not feasible for large cohort studies. Furthermore, it is subject to human error in data analysis.

Computational phenotyping aims to automatically mine or predict clinically significant, or scientifically meaningful, phenotypes from structured EHR data, unstructured clinical narratives, or combination of the two. In this study, we aim to develop a model to identify distant recurrences within a cohort of breast cancer patients. To develop the model, we utilize data collected in Northwestern Medicine Enterprise Data Warehouse (NMEDW), which is a joint initiative across the Northwestern University Feinberg School of Medicine and Northwestern Memorial HealthCare [5]. The NMEDW houses the EHR for about 6 million patients. Both structured and unstructured data are available in the NMEDW. Structured data typically capture patients' demographic information, lab values, medications, diagnoses, and encounters [6]. Although readily available and easily accessible, studies have concluded that structured data alone are not sufficient to accurately infer phenotypes [7, 8]. For example, ICD-9 codes are mainly recorded for administrative purposes and are influenced by billing requirements and avoidance of liability [9, 10]. Consequently, these codes do not always accurately reflect a patient's underlying physiology. Furthermore, not all patient information (such as clinicians' observations and insights) is well documented in structured data [11]. As a result, using structured data alone for phenotype identification often results in low performance [8]. The limitations associated with structured data for computational phenotyping have encouraged the use of clinical narratives, which typically include clinicians' notes, observations, referring letters, specialists' reports, discharge summaries, and records of communication between doctors and patients [12]. These clinical narratives contain rich descriptions of patients' disease assessment, history, and treatments. However, the clinical narratives are not readily accessible without the use of natural language processing (NLP). The abundance of information in the free text makes NLP an indispensable tool for text-mining [13-15].

Our goal is to develop such a system that combines structured EHR data and unstructured clinical narratives to accurately and efficiently identify distant recurrences in breast cancer. Such a model can be easily replicated and requires a minimum amount of human effort and input.

## II. RELATED WORK

Computational phenotyping has facilitated biomedical and clinical research across many applications, including patient diagnosis categorization, novel phenotype discovery, clinical trial screening, pharmacogenomics, drug-drug interaction (DDI) and adverse drug event (ADE) detection, and downstream genomics studies. Different NLP applications have also been developed to identify breast cancer recurrences. Carrell et al. [16] proposed a method to identify breast cancer sub-cohorts with ipsilateral, regional, and metastatic events using the concepts identified within the free text. The binary classification model achieved an F-measure scores of 0.84 and 0.82 in the training set and test set, respectively. However, the model could not distinguish a local recurrence from a distant recurrence. In addition, defining the number of hits in the system to segment the documents required substantial effort. Using morphology codes and anatomical sites from pathology reports, Strauss et al. [17] were able to identify recurrences. However, their approach required that the pathology reports be well-documented under a standard format. However, the majority of distant recurrences in breast cancer have been diagnosed clinically rather than pathologically [18]. It has been challenging to identify distant recurrences from pathology reports because they are not usually recorded as clinical diagnoses in the reports. Haque et al. [19] applied a hybrid approach to identify breast cancer recurrences using a combination of pathology reports and EHR data. They achieved a relatively high NPV of 0.995 and a relatively low PPV of 0.65. This model also required a minimum amount of ten percent manual chart review, which is still fairly time-consuming. In addition, the model was not able to distinguish between local, regional, or distant recurrences. NLP has also been applied to attempt retrieving distant recurrences for other types of cancer. Lauren et al. [20] tried to identify distant recurrences in prostate cancer from clinical notes, radiology reports, and pathology reports. They concluded that NLP could be used to identify metastatic prostate events more accurately than claim data.

Clinical narratives are known to have high-dimensional feature spaces, few irrelevant features, and sparse instance vectors [21]. These problems were found to be well-addressed by SVMs [21], which also have been recognized for their generalizability and are widely used for computational phenotyping [13, 22-27]. Carroll et al. [26] implemented a SVM model for rheumatoid arthritis identification using a set of features from clinical narratives using the Knowledge Map Concept Identifier (KMCI) [28]. They demonstrated that a SVM algorithm trained on these features outperformed a deterministic algorithm.

A combination of structured data and narratives for phenotyping have been found to improve model performances. DeLisle et al. [29] implemented a model to identify acute respiratory infections. They used structured data combined with narrative reports and demonstrated that the inclusion of free text improved the PPV score by 0.2–0.7 while retaining sensitivities around 0.58-0.75. In a study of the identification of methotrexate-induced liver toxicity in patients with rheumatoid arthritis, Lin et al. [30] obtained an F-measure of 0.83 in a performance evaluation. Liao et al. [31] implemented a penalized logistic regression as a classification algorithm to predict patients' probabilities of having Crohn's disease and achieved a PPV score of 0.98. Both Lin's and Liao's methods experimented with a combination of features from structured EHR and NLP-processed features from clinical narratives. Their studies showed that the inclusion of NLP methods resulted in significantly improved performance.

## III. METHODOLOGY

### A. Cohort Description

TABLE 1: COHORT DISTRIBUTION IN THE TRAINING AND GENERALIZATION SET

|  | Total | Distant Recurrence | Percentage (%) | Overall percentage (%) |
|---|---|---|---|---|
| **Double-annotated set** | 1,995 | 193 | 9.87% | |
| Cross-validation set | 1,396 | 138 | 9.89% | 9.22% |
| Held-out test set | 599 | 55 | 9.19% | |
| **Single-annotated set** | 4,904 | 443 | 9.03 % | |

Patients diagnosed with breast cancer between 01/01/2001 and 12/31/2015 are drawn from NMEDW. Patients are identified by ICD-9 codes. In total, 19,874 females are included. Within this cohort, only cases with at least one surgical pathology report documented in the desired time window are selected. In total, 6,899 subjects are identified and included in this study. The workflow to generate this data set is presented in Figure 1.

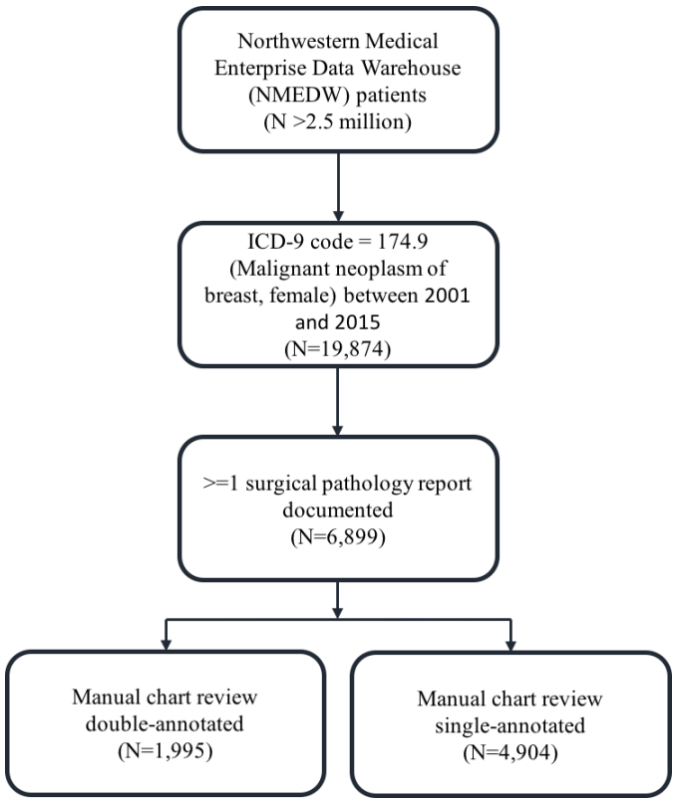

Figure 1: Workflow to identify the cohort

To establish a gold standard for algorithm development, each patient is assigned a definite distant recurrence status ('Yes' or 'No') according to manual chart review. In total, 1,995 subjects are annotated twice by two annotators (co-authors: medical student AR; breast surgery fellow SE) and are included for model training and validation. The inter-rater agreements for the two annotators are measured by Cohen's kappa score, and the obtained score is 0.87 [32]. The items without agreements are resolved by a discussion between the two annotators. The other 4,904 subjects are annotated once by annotators (co-authors: post-doc fellow XL; Ph.D. candidate ZZ) and are used as an independent set for model generalization test. These annotations are conducted over a span of 15 months (completed September 2017).

The 1,995 double-annotated subjects are randomly split into a cross-validation set and a held-out test set according to a 7:3 ratio. In the cross-validation set, five-fold cross-validation is applied with the 1,396 samples. Among these 1,396 samples, 138 distant recurrence events are identified; among the 599 samples in held-out test set, 55 distant recurrences are identified. In the generalization test set, 443 distant recurrences are identified among the 4,904 samples. The cohort distribution is shown in TABLE 1.

*B. Structured Clinical Data*

Automated SQL codes are developed to query structured data from NMEDW. In total, 18 structured clinical variables are retrieved or derived. The variable names and corresponding categories or values are displayed in TABLE 2. Demographic data such as the age of diagnosis, race, smoking history, alcohol usage, family cancer history, and insurance type are queried. Smoking history is categorized as 'Yes', 'No', 'Ex-smoker', or 'Unknown'. Alcohol usage is categorized as 'No', 'Moderate', 'Heavy', 'Former', or 'Unknown'. Tumor characteristics and biomarkers, such as estrogen receptor (ER), progesterone receptor (PR), HER2, P53, nodal positivity, histology, tumor grade, and tumor size are retrieved. Nodal positivity is categorized as 'Positive', 'Negative', or 'Unknown'. The variable histology and nodal positivity are selected, because subjects with invasive ductal breast cancer or positive lymph nodes are more likely to develop a distant recurrence compared to those that have ductal in situ or negative lymph nodes [33]. Primary surgery type is categorized as 'Breast conservation surgery', 'Mastectomy', 'No', or 'Unknown'.

Additional clinical variables are derived to help identify distant recurrences. Variables of deceased, targeted therapy, and radiation are developed. The deceased variable is a binary variable to indicate whether a patient deceased before the age of 75. Intuitively, patients with distant recurrences might have a shorter survival length compared to the women who do not have distant recurrences. After a discussion with a domain expert (co-author SK), we choose the age of 75 as the cutoff. Another variable 'targeted therapy' is a binary variable created to indicate whether the patient has taken any of the following drugs: 'Afinitor', 'Everolimus', 'Bevacizumab', 'Avastin',

'Ibrance', or 'Palbociclib'. These drugs are prescriptions for patients with distant recurrences. An additional variable radiation is a binary variable indicating whether the subject has received radiation treatment at the site of metastases, such as brain, lung, or bone. This variable is derived from the intuition that patients receiving radiation at a site different from the primary tumor are at a higher chance of having distant recurrences.

TABLE 2: THE NAME AND CORRESPONDING CATEGORIES (VALUES) OF THE 18 RETRIEVED STRUCTURED CLINICAL VARIABLES. IDC IS INVASIVE DUCTAL CARCINOMA, DCIS IS DUCTAL CARCINOMA IN SITU, ILC IS INVASIVE LOBULAR CARCINOMA, NETWORK CATEGORY IS THE NETWORK OF PATIENT'S INSURANCE PLAN.

| Variable Name | Category |
|---|---|
| age of diagnosis | Continuous |
| race | White, Black, Asian, Other |
| smoking history | Yes, No, Ex-smoker, Unknown |
| alcohol usage | No, Moderate, Heavy, Former, Unknown |
| family cancer history | Yes, No, Unknown |
| insurance type | Network Category |
| estrogen receptor | Positive, Negative, Unknown |
| progesterone receptor | Positive, Negative, Unknown |
| HER2 | Positive, Negative, Unknown |
| P53 | Positive, Negative, Unknown |
| nodal positivity | Positive, Negative, or Unknown |
| histology | IDC, DCIS, ILC, Unknown |
| grade | Grade1, Grade2, Grade3, Unknown |
| size | 0-2cm, 2cm-5cm, >5cm, Unknown |
| surgery type | Mastectomy, Breast conservation surgery, Unknown |
| deceased | Yes, No |
| targeted therapy | Yes, No |
| radiation | Yes, No |

### C. Clinical Narratives

We query the NMEDW for clinical narratives generated before May 2016 (the start time of manual chart review) or the date when the patient is censored. All inpatient and outpatient notes are retrieved without any provider type restriction. The retrieved clinical narratives include progress notes, pathology reports, telephone encounter notes, assessment and plan notes, problem overview notes, treatment summary notes, radiology notes, lab notes, procedural notes, and nursing notes. Only notes generated after the diagnosis of breast cancer are retrieved. We only include the notes having at least one mention of 'breast'. After retrieving the narratives, we first preprocess the corpus by removing duplicate copies, tokenizing sentences, and removing non-English symbols. Following these preprocessing steps, we annotate the medical concepts in the sentences using MetaMap, an NLP application to map the biomedical text to the UMLS Metathesaureus [34]. The surrounding semantic context is determined. CUIs that are tagged as negated by NegEx [35] are excluded (NegEx is a negation tool configured in MetaMap). If multiple CUIs are mapped, the one with maximum MMI score (a score ranked by MetaMap) is retained. In order to completely and accurately exclude negations or unrelated contextual cues, such as a differential diagnosis event, sentences with negative contextual features (e.g., 'no', 'rule out', 'deny', 'unremarkable') and uncertain contextual features (e.g., 'risk', 'concern', 'worry', 'evaluation') are also removed. This customized list of contextual features is obtained from the development corpus.

### D. Feature Generation

To focus our NLP efforts, we identify a set of target distant recurrence concepts with the help of sample notes. We review a development corpus of ten randomly selected samples' notes with distant recurrences and extract partial sentences that are related to a breast cancer distant recurrence. These extracted partial sentences appear in TABLE S1. The initial set contains 20 partial sentences. These partial sentences are tagged by MetaMap, and the CUIs corresponding to each concept is obtained. The customized dictionary contains 83 CUIs (TABLE S2). After data preprocessing and concept mapping, only CUIs with highest MMI score that also fall within the customized dictionary are used as features for model training. CUIs with MMI score smaller than one are filtered and excluded. Following this feature selection, there are 83 narrative-based features remaining for inclusion in the machine learning algorithm. In addition to the obtained CUI features, the 18 structured clinical variables described above are used as additional features.

### E. Prediction Model and Evaluation

We use support vector machine (SVM) to develop an algorithm to predict whether patients had distant recurrences. SVMs have been widely used for computational phenotyping [13, 22-27]. We apply linear kernel type for the SVM models. In our experiments, we train four baseline classifiers on different feature types: a full set of medical concepts tagged by MetaMap [34], a filtered set of medical concepts tagged by MetaMap, only the structured clinical data, and a standard bag of words from clinical narratives. TfIDFVectorizer class in scikit-learn is used to convert the raw documents to a matrix of TF-IDF features to assemble a bag of words. In the full MetaMap and bag of words, Chi-square test is applied to select features before training the model to remove the common words that exist in clinical narratives. Only top 5% features are retained for modeling.

In the model evaluation, we choose area under curve (AUC) score as a measurement metric because this is a skewed cohort with low event rate. The output of our SVM model is probabilities, though in practice, various thresholds result in different true positive/false positive rates and AUC score considers all possible thresholds. To better demonstrate the thresholds and model performance, the corresponding receiver operating characteristic (ROC) curves for the different methods are evaluated. Cross validation performance depends on the randomly shuffled split of the training dataset into multiple folds. In order to obtain robust performance statistics, each five-fold cross validation is replicated 20 times using shuffled stratified splits initialized with different random seeds.

TABLE 3: DESCRIPTIVE SUMMARIES OF 1,995 SUBJECTS' CLINICAL DATA. THE SIGNIFICANCE TEST IS PERFORMED BETWEEN THE RECURRENCE GROUP AND THE NON-RECURRENCE GROUP. ONLY DATA WITH P-VALUES LESS THAN 0.05 ARE PRESENTED. DR STANDS FOR DISTANT RECURRENCE. THE MEAN AND STANDARD DEVIATION ARE CALCULATED FOR CONTINUOUS VARIABLES. NUMBERS AND PERCENTAGES ARE PRESENTED FOR CATEGORICAL VARIABLES. P-VALUES ARE OBTAINED USING STUDENT'S T-TEST FOR CONTINUOUS VARIABLES AND CHI-SQUARED TEST FOR CATEGORICAL VARIABLES.

|  | Double-annotated set N=1,995 | DR N=193 | No DR N=1,802 | P-value |
|---|---|---|---|---|
| **Nodal positivity (%)** | 544 (27.3%) | 103 (53.4%) | 441 (24.5%) | 1.4e-14 |
| **Histology (%)** |  |  |  | 2.6e-06 |
| IDC | 1,530 (76.7%) | 174 (90.2%) | 1,356 (75.2%) |  |
| DCIS | 279 (14.0%) | 3 (1.6%) | 276 (15.3%) |  |
| ILC | 155 (7.8%) | 15 (7.8%) | 140 (7.8%) |  |
| **Grade (%)** |  |  |  | 2.1e-10 |
| Grade 1 | 458 (23.0%) | 16 (8.3%) | 442 (24.5%) |  |
| Grade 2 | 851 (42.7%) | 73 (37.8%) | 778 (43.2%) |  |
| Grade 3 | 665 (33.3%) | 101 (52.3%) | 564 (31.3%) |  |
| **Deceased (%)** | 157 (7.9%) | 98 (50.8%) | 59 (3.3%) | < 2.2e-16 |
| **Radiation (%)** | 67 (3.4%) | 52 (26.9%) | 15 (0.8%) | < 2.2e-16 |
| **Targeted therapy (%)** | 60 (3.0%) | 44 (22.8%) | 16 (0.9%) | < 2.2e-16 |

## IV. EXPERIMENT RESULTS

As demonstrated in TABLE 3, clinical data with a significant difference between the recurrence group and the non-recurrence group in the double-annotated training set are presented. Compared to the non-recurrence patients, women with recurrences had a higher percentage of nodal positivity and higher grade of tumor, were more likely to be diagnosed with invasive ductal carcinomas, had more radiation performed at the metastasis site, had received more targeted therapies, and were more likely to die before the age of 75.

Using SVM as a prediction model, the AUC scores obtained from the cross-validation are reported in TABLE 4. To note, the model applied was an SVM model with linear kernel (C equaled 1, and gamma was set as 'auto' in the python package 'sklearn.svm'). The AUC score obtained in our proposed model was 0.92 (SD=0.01). The performance of our proposed model significantly outperformed the other four baselines. The P-value for Student's t-test was 0.0004 comparing our proposed model with the second-ranked model of Filtered MetaMap.

To illustrate, the corresponding receiver operating characteristic (ROC) curves for the different methods are plotted in Figure 2.

TABLE 4: THE NUMBER OF FEATURES AND THE AUC SCORES OBTAINED IN THE CROSS-VALIDATION USING 70% OF THE GROUND TRUTH DATA.

| Model | Number of Features | AUC (SD) |
|---|---|---|
| Filtered MetaMap +Clinical Data | 101 | 0.92 (0.01) |
| Full MetaMap | 1,537 | 0.78 (0.04) |
| Filtered MetaMap | 83 | 0.90 (0.02) |
| Clinical Data | 18 | 0.77 (0.04) |
| Bag of Words | 4,959 | 0.82 (0.02) |

We trained an SVM model on the training set (1,396 samples) and then predicted labels on the held-out test set (599 samples). Comparing the predicted probabilities and the annotated labels, the obtained AUC scores are presented in TABLE 5. The AUC score obtained in our proposed model was 0.95. The model with NLP-features, Filtered MetaMap also had a notable performance of 0.93. The performance in our proposed model again outperformed all the baseline models.

TABLE 5: THE NUMBER OF FEATURES AND THE AUC SCORES OBTAINED IN THE EXTERNAL TEST USING THE TEST SET (599 SAMPLES).

| Model | AUC |
|---|---|
| Filtered MetaMap + Clinical Data | 0.95 |
| Full MetaMap | 0.56 |
| Filtered MetaMap | 0.93 |
| Clinical Data | 0.87 |
| Bag of Words | 0.55 |

In addition to our training and validation analyses, we applied our fitted model to predict labels on the generalization set, which contained 4,904 single-annotated samples. In this generalization test, we obtained an AUC score of 0.93, which had a similar performance as the held-out test.

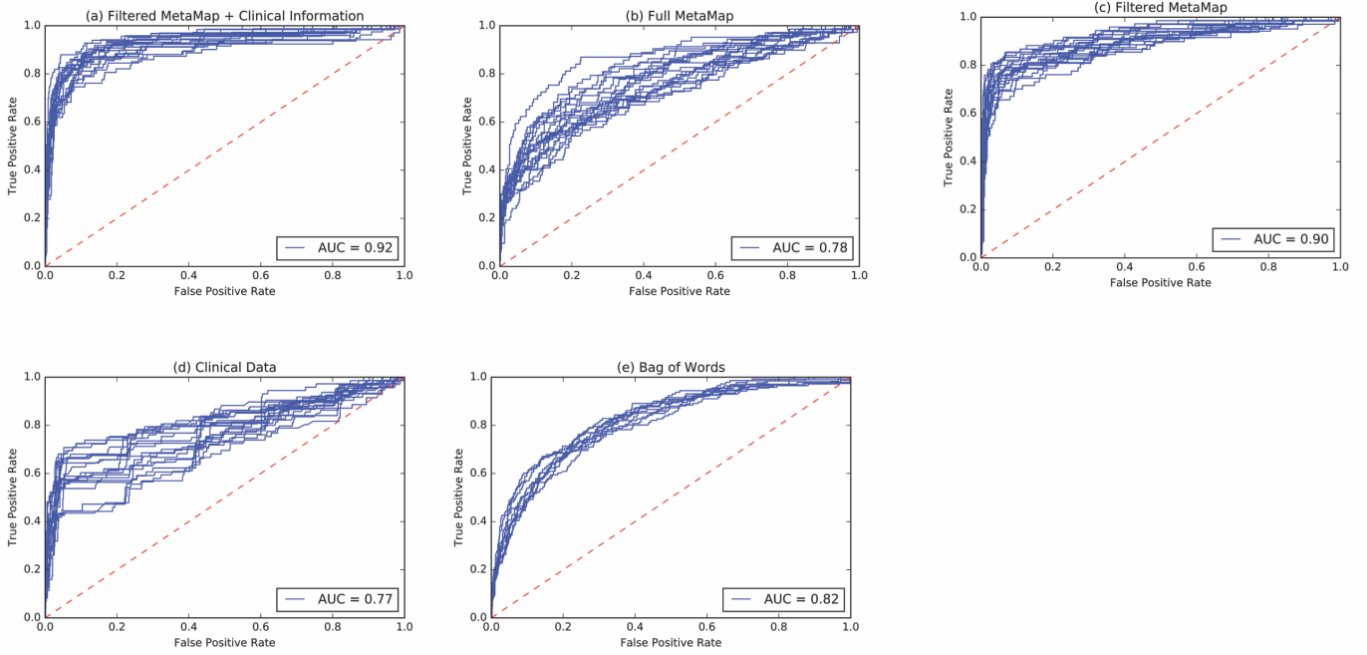

Figure 2: Receiver operating characteristic (ROC) curves using different methods. Each experiment is replicated 20 times using different shuffled stratified splits and all derived ROC curves are plotted.

TABLE 6: TOP 15 VARIABLES WITH THEIR CORRESPONDING COEFFICIENTS

| CUI | Name | Coefficient | Partial Sentences |
|---|---|---|---|
| C0153678 | Secondary malignant neoplasm of pleura | 1.00 | cancer metastatic to pleura metastatic cancer to pleura |
| Radiation | Clinical Variable | 0.90 | |
| Deceased | Clinical Variable | 0.90 | |
| Targeted therapy | Clinical Variable | 0.84 | |
| C0153690 | Secondary malignant neoplasm of bone | 0.76 | metastases to bone, bone metastases |
| C1967552 | IXEMPRA | 0.71 | ixempra |
| C0278488 | Carcinoma breast stage IV | 0.70 | metastatic breast cancer, breast cancer stage iv, metastatic breast carcinoma |
| C0494165 | Secondary malignant neoplasm of liver | 0.62 | liver metastases, liver metastatic disease, metastatic disease liver, metastases to the liver, liver metastases |
| C0220650 | Metastatic malignant neoplasm to brain | 0.59 | brain metastases |
| C1266909 | Entire bony skeleton | 0.39 | bone |
| C2939420 | Metastatic Neoplasm | 0.27 | metastatic disease |
| C0036525 | Metastatic to | 0.25 | metastatic |
| C0027627 | Neoplasm Metastasis | 0.25 | metastatic disease |
| C0346993 | Secondary malignant neoplasm of female breast | 0.23 | metastatic breast cancer to the |
| C1522484 | Metastatic qualifier | 0.22 | metastatic |

From the fitted SVM model using the 1,396 samples in the training set, we retrieved the coefficient scores for each feature. The top 15 ranked coefficient scores and their corresponding variable names appear in TABLE 6. Three of the clinical variables (radiation, deceased, and targeted therapy) were highly ranked on the list. These three variables were treatment or outcome variables. The rest of the top-ranked features were concepts obtained from clinical narratives. Most of the CUIs were either related to metastases events or related to the metastatic sites that breast cancer could spread to. The term 'IXEMPRA' is a prescription medicine used for locally advanced breast cancer or breast cancer with distant recurrences.

## V. Discussion

In this study, we combined 83 features from unstructured clinical narratives and 18 features from structured clinical data to identify distant recurrences in breast cancer. Clinical narratives were extracted from progress notes, pathology reports, telephone encounter notes, assessment and plan notes, problem overview notes, treatment summary notes, radiology notes, lab notes, procedural notes, and nursing notes generated after diagnosis of primary breast cancer. The clinical narratives were tagged by NLP application MetaMap to generate UMLS concepts. After filtering out concepts that were not in the customized dictionary, the remaining concepts were combined with the structured clinical data to train an SVM model for distant recurrence identification. We were able to identify structured clinical variables that could stratify the groups of women with and without distant recurrences. Using such a method, we obtained an AUC score of 0.95 and 0.93 in our external held-out test and generalization test.

During the feature coefficient study, we found that the features "secondary malignant neoplasm of pleura, radiation, deceased, targeted therapy, and secondary malignant neoplasm of bone" were the top-ranked features. Intuitively, women with distant recurrences have a higher chance of receiving radiation at the metastatic site and of receiving targeted therapy compared to those without distant recurrences. They are also more likely to have a lower survival rate. The most common sites of metastasis to organs were the bone, brain, lung, and liver [1]. In our study, we found the mentions of metastatic to bone, liver, and brain were also top-ranked. The terms 'metastatic' and 'breast cancer' were also more likely to appear in the clinical notes of patients with distant recurrences.

Progress notes are notably telegraphic. Also, many excessively busy residents and senior clinicians create notes by simply copying and pasting previous encounter notes, while making only minor updates for the most recent appointment. This results in many notes that differ in 'critical' content still scoring highly on the overall measures of similarity. The same applies to the full set of MetaMap concepts, which is similar to the bag of words. To include only highly associated features in our model, we removed the common concepts or words in the notes. Chi-square test was applied to select features before training the model. Only the top 5% features were retained for full MetaMap and bag of words modeling. This test might have the potential for overfitting in cross-validations. Indeed, we saw a lower performance in the held-out test for full MetaMap and bag of words. To adjust this problem, we tested different thresholds for the Chi-square test selection. However, we found 5% ended with the best results.

Identifying breast cancer distant recurrence in clinical data sets is important for clinical research and practice. Annotation of distant recurrence is difficult using standard EHR phenotyping approaches and are commonly beyond the scope of manual annotation efforts by cancer registries. A model using natural language processing, EHR data, and machine learning to identify distant recurrences in breast cancer patients allows more accurate data-mining and significantly less time-consuming manual chart review. We expect that by minimally adapting the positive concept set, this study has the potential to be replicated at other institutions with a moderately sized training dataset. In this study, we generated features using sentences extracted from the clinical narratives combined with structured data. The training and testing data sets were cross-annotated in the process, which offered a solid ground truth for the study. Replicating this model requires minimal outside effort. We offered the customized dictionary in this study, so a user can retrieve the required notes and clinical structured data in order to replicate this study. After the rigorous manual chart review and feature retrieval, our data set has offered a gold-standard data set with rich, validated information for further breast cancer research.

When replicating this study at another institution, there is a chance that one will not be able to find the structured clinical data in their databases. If this is the case, some of the structured data can be found from other resources. Variables of 'histology' and 'lymph node status' can be extracted from pathology reports using a rule-based system. For example, expressions of 'total lymph nodes', 'total lymph nodes number positive', 'axillary lymph nodes examined', 'axillary lymph nodes examined number of positive versus total' can be used to extract lymph node status from pathology report at our institution. Survival information can be found in the administrative billing system.

## VI. Future Work

The NLP pipeline cannot characterize the context of features. Clinical narratives contain patients' concerns, clinicians' assumptions, and patients' past medical histories. Clinicians also record diagnoses that are ruled out or symptoms that patients denied. Our next aim will be that such conditions, mentions, and feature relations will be extracted to better distinguish differential diagnoses. Generalized relation and event extraction, rather than binary relation classification, will be conducted. To this end, graph methods are a promising class of algorithms and should be actively investigated [36, 37].

In the future, we will test our data with different machine learning models. In this study, we have chosen SVM model with linear kernel for interpretation purposes. Other models might result in better performance.

We will also aim to address the heterogeneity problem in clinical narratives. It is a common problem in clinical narratives due to the variance in physicians' expertise and behaviors [38]. Features derived from clinical narratives included in this study were extracted from notes generated by clinicians with different specialties and professional levels of expertise. As a result, some content was not relevant to the breast cancer distant recurrence event, even though we had limited the notes to include the mention of 'breast'. For example, a liver cancer metastasis to the breast from a primary liver tumor would be difficult to identify. We will need to resolve the heterogeneity in clinical narratives.

## VII. CONCLUSIONS

We developed a machine learning model by combining structured clinical data and unstructured clinical narratives in order to identify distant recurrence events in breast cancer. We demonstrated the high accuracy and efficiency of our model, using cross-validation, held-out test evaluation, and a further generalization set evaluation. Our proposed model allows for more accurate and efficient identification of distant recurrences than single modality models using either clinical narratives or structured clinical data. Thus, our model is a significantly less time-consuming and practical alternative to manual chart review. This is particularly relevant in an era when evidence-based medicine receives growing attention and there is more emphasis on computational phenotyping and data-driven discovery. This model would also be valuable and applicable to research in other medical fields beyond breast cancer.

## ACKNOWLEDGMENT

This project is supported in part by NIH grant R21LM012618-01.